\documentclass[letterpaper]{article}
\usepackage[preprint]{aaai2027}
\usepackage[hyphens]{url}
\usepackage{graphicx}
\usepackage{natbib}
\usepackage{caption}
\usepackage{booktabs}
\usepackage{amsmath}
\usepackage{amsfonts}
\usepackage{amssymb}
\usepackage{mathtools}
\usepackage{placeins}
\usepackage{algorithm}
\usepackage{algorithmic}
\usepackage{tabularx}
\usepackage{multirow}

\DeclareCaptionStyle{ruled}{labelfont=normalfont,labelsep=colon,strut=off}
\newcommand{\coderepolink}[1]{\leavevmode\pdfstartlink attr {/Border [0 0 0]} user {/Subtype /Link /A << /S /URI /URI (#1) >>}\url{#1}\pdfendlink}
\newcommand{\scoreup}[1]{\textcolor{green!50!black}{\scriptstyle #1}}
\newcommand{\scoredown}[1]{\textcolor{red}{\scriptstyle #1}}
\newcommand{\scorezero}[1]{\textcolor{green!50!black}{\scriptstyle #1}}
\newcommand{\citepboxed}[1]{\citep{#1}}

\title{BiVerify: Closing the Loop Between Backward and Forward Verification for Mathematical Self-Correction}
\author{
    Rui Zou\textsuperscript{\rm 1},
    Xin Zhao\textsuperscript{\rm 1},
    Mengqi Wei\textsuperscript{\rm 2},
    Ji-Rong Wen\textsuperscript{\rm 1}\corresponding
}
\affiliations{
    \textsuperscript{\rm 1}Gaoling School of Artificial Intelligence, Renmin University of China\\
    \textsuperscript{\rm 2}National Engineering Research Center of Educational Big Data, Central China Normal University\\
    zouruixyz@ruc.edu.cn, batmanfly@gmail.com, weimengqi@mails.ccnu.edu.cn, jrwen@ruc.edu.cn
}
\graphicspath{{Figures/BiVerify/}}

\makeatletter
\g@addto@macro\@maketitle{%
  \g@addto@macro\@thanks{%
    \footnotetext[0]{\textbf{Code}\enspace\coderepolink{https://github.com/TicusFFF/BiVerify}}}}
\makeatother
\begin{document}
\maketitle

\begin{abstract}
Large language models (LLMs) have made substantial progress in mathematical reasoning but still produce incorrect solutions to complex problems, indicating the need for more reliable self-correction mechanisms. Mathematical self-correction typically identifies errors through verification, which can proceed backward or forward. The two directions provide complementary verification capabilities. Backward verification checks a candidate answer against the original constraints, bypassing complex intermediate reasoning, but cannot ensure correct reasoning or objective completeness. Forward verification follows the problem conditions and current solution, enabling it to locate reasoning errors and assess objective completeness, but may overlook key constraints in complex solution chains. We propose BiVerify, a dynamically scheduled framework for bidirectional mathematical verification. BiVerify begins with backward verification to screen the answer against the original constraints, followed by forward verification of the reasoning process; subsequent direction switching, state resetting, and re-verification are governed by verification outcomes. Upon failure, the system revises the candidate or verification flow until the same candidate passes both directions. We also introduce shared mathematical tools that separate logical reasoning from precise computation. We systematically evaluate BiVerify on five challenging mathematical reasoning benchmarks using multiple mainstream DeepSeek, GPT, and Gemini models. BiVerify exceeds the strongest representative public baseline evaluated in this work by \(13.5\) percentage points in average accuracy, with the margin reaching \(31.2\) percentage points on GPT. Further analysis shows that BiVerify's overall performance exceeds that of either verification direction alone.
\end{abstract}
 \section{Introduction}

\begin{figure}[t]
    \centering
    \includegraphics[width=\linewidth]{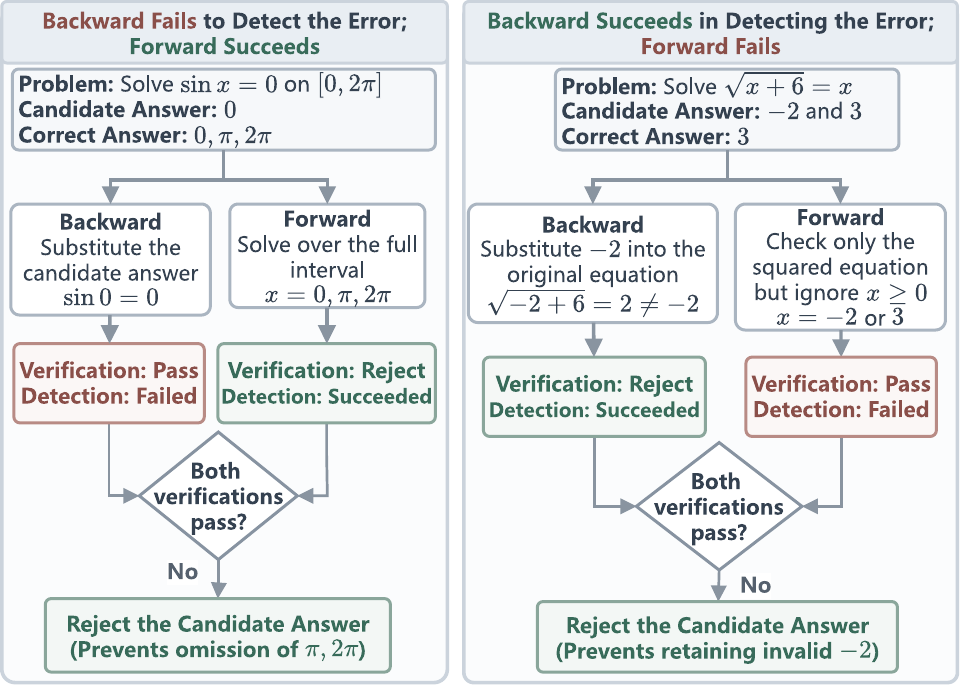}
    \caption{
    Complementarity between backward and forward verification in detecting mathematical reasoning errors.
    }
    \label{fig:biverify_motivation}
\end{figure}

Large language models (LLMs) have made substantial progress in mathematical reasoning~\citepboxed{deepseekai2025r1,luo2025wizardmath,yu2025chain}, yet they still err on complex problems~\citepboxed{huang2024selfcorrect,tyen2024llms,yuan2026miraclesteps}. Strengthening their ability to identify and correct such errors is therefore an important research direction. Existing self-correction approaches revise LLM outputs through iterative model-generated feedback in Self-Refine~\citepboxed{madaan2023selfrefine}, natural-language reflection in Reflexion~\citepboxed{shinn2023reflexion}, checklist-guided evaluation in TICK~\citepboxed{cook2024ticking}, and program-driven verification and refinement in ProgCo~\citepboxed{song-etal-2025-progco}. Although these approaches construct feedback differently, correction still relies on verifying whether the current solution is correct.

Mathematical verification can proceed in two directions. Backward verification starts from the candidate answer and checks it against the original problem constraints. ProgCo provides a representative example by using program-driven checks of a candidate response and execution feedback to guide refinement~\citepboxed{song-etal-2025-progco}. This direction can bypass lengthy or potentially flawed intermediate reasoning, but passing the checked constraints does not guarantee that the reasoning process is correct or that the solution objective is complete. Forward verification instead follows the problem conditions and the current solution, as exemplified by step-level process feedback~\citepboxed{lightman2024verify}. It can examine intermediate reasoning, localize process-level errors, and assess objective completeness. However, when the solution chain is long or the verification flow is incomplete, forward verification may itself become error-prone and miss key constraints. The two directions are therefore complementary: forward verification examines intermediate reasoning and objective completeness, whereas backward verification independently checks the answer while bypassing complex intermediate processes. Figure~\ref{fig:biverify_motivation} illustrates this complementarity through two examples.

This complementarity suggests a natural verification order: backward verification first checks the candidate against the original constraints, after which forward verification examines the reasoning process. Beyond this order, prior work has paid less attention to how backward and forward verification states should be coordinated as the same candidate is iteratively revised. Revising a candidate invalidates the verification results obtained for its previous version, whereas an inadequate verification flow should be repaired without unnecessarily changing the candidate. The system must therefore distinguish between candidate revision and verification-flow revision, re-check the affected direction, and accept a result only when both directions pass on the same candidate.

We therefore propose BiVerify, a dynamically scheduled framework for bidirectional mathematical verification. Its scheduler maintains the current candidate and both verification states, coordinating direction switching, state resetting, and re-verification in a closed loop. Because LLMs may still err in exact computation, BiVerify provides shared mathematical tools for both verification directions. The LLM constructs verification objectives and organizes the reasoning, while the shared tools perform precise symbolic and numerical operations, thereby decoupling reasoning from execution. Tool outputs serve only as verification evidence; the adjudication agents make the directional pass/fail decisions, while the scheduler determines final acceptance.

Across five challenging mathematical reasoning benchmarks and seven DeepSeek, GPT, and Gemini settings, BiVerify improves seven-model average accuracy by \(13.5\) percentage points over the strongest evaluated public baseline; this margin reaches \(31.2\) percentage points in the corresponding GPT setting. Ablations further show that BiVerify corrects more erroneous answers than either direction alone without materially increasing errors on correct answers, yielding stronger overall performance.

Our contributions are as follows: (1) We identify complementary roles for backward verification from a candidate answer and forward verification from the solution process: the former bypasses complex intermediate reasoning to check the answer against the original constraints, whereas the latter verifies ordered key claims, the reasoning process, and objective completeness; (2) We propose BiVerify, which dynamically schedules both directions in a closed loop and requires both to pass on the same candidate; and (3) Experiments on five challenging mathematical reasoning benchmarks and seven base-model settings demonstrate BiVerify's advantages over the evaluated public baselines and either verification direction alone.
 \section{Related Work}
\label{sec:related_work}

Three areas are most relevant to this work: LLM self-correction, mathematical-reasoning verification, and tool-augmented mathematical reasoning, which improve reliability through feedback, verification targets, and external evidence, respectively.

\noindent\textbf{Test-Time Improvement and Self-Correction of LLMs.} LLM self-correction improves inference-time outputs through feedback, reflection, or repeated sampling. Self-Refine iteratively revises outputs using self-generated feedback~\citepboxed{madaan2023selfrefine}; Reflexion converts task feedback into linguistic reflections~\citepboxed{shinn2023reflexion}; TICK structures evaluation with generated checklists~\citepboxed{cook2024ticking}; and Self-Consistency selects consensus across sampled reasoning paths~\citepboxed{wang2022selfconsistency}. Pan et al. survey automatic correction strategies~\citepboxed{pan2024automatically}, while Kamoi et al. summarize conditions under which models can correct their errors~\citepboxed{kamoi2024selfcorrection}. Huang et al. find that models may not improve reliably without dependable external feedback~\citepboxed{huang2024selfcorrect}; Tyen et al. show that LLMs often fail to locate reasoning errors but can correct them when the error location is provided~\citepboxed{tyen2024llms}.

\noindent\textbf{Verification in Mathematical Reasoning.} Prior work targets either candidate outputs or reasoning processes. Cobbe et al. train verifiers to score and rank multiple outputs and select more reliable results~\citepboxed{cobbe2021gsm8k}, but this selection paradigm differs from our backward verification because it does not reason from a candidate answer back to the original constraints. More directly related, ProgCo reasons backward from an output answer to detect contradictions and uses program-execution results for refinement~\citepboxed{song-etal-2025-progco}. For process verification, Lightman et al. use step-level supervision to provide finer-grained reasoning signals~\citepboxed{lightman2024verify}, while ProcessBench evaluates process verifiers with first-error localization as a central target~\citepboxed{zheng2025processbench}.

\noindent\textbf{Tool-Augmented Mathematical Reasoning and Agents.} Tool-augmented reasoning uses programs, interpreters, or specialized tools to improve exact computation and symbolic execution. PAL translates problems into executable Python programs~\citepboxed{gao2023pal}; Program-of-Thoughts separates computation from natural-language reasoning~\citepboxed{chen2023pot}; and ToRA alternates between reasoning and external tools~\citepboxed{gou2024tora}. For tool use and orchestration, ReAct interleaves reasoning with external actions~\citepboxed{yao2023react}, while Toolformer learns when and how to call external APIs~\citepboxed{schick2023toolformer}. AgentMath integrates tools into mathematical reasoning agents~\citepboxed{luo2025agentmath}, and R1-Code-Interpreter supports computation and verification in complex problems through a code interpreter~\citepboxed{chen2025r1codeinterpreter}.

Prior work has advanced self-correction, candidate scoring, directional verification, and tool-augmented reasoning, but has paid less attention to coordinating backward and forward verification states as the same candidate is revised. BiVerify organizes both directions and shared tool evidence into a closed loop over that candidate.
 \section{Method}

\begin{figure*}[t]
    \centering
    \includegraphics[width=0.95\textwidth]{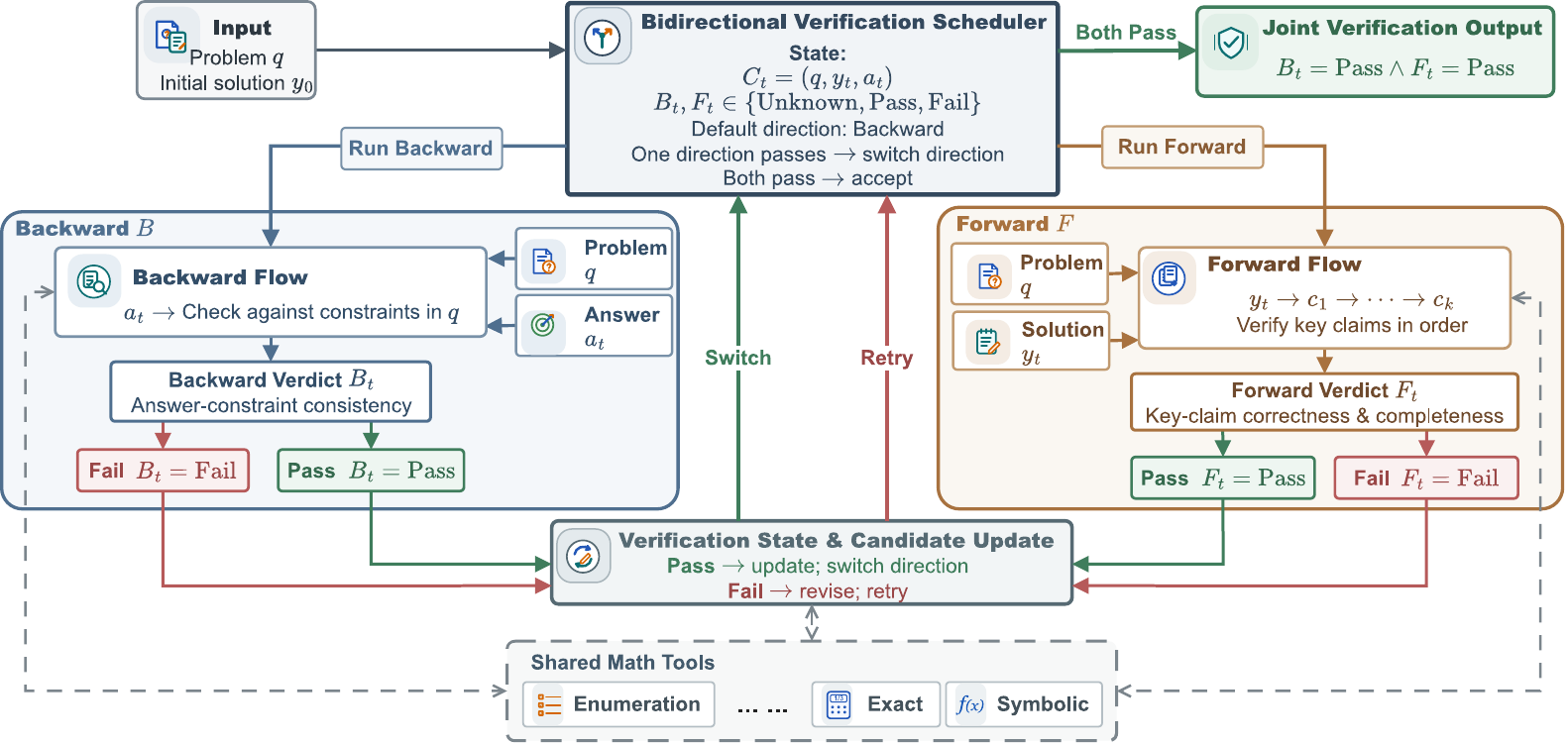}
    \caption{
    Overview of BiVerify. The system performs backward and forward verification on the same candidate, while the scheduler coordinates direction switching, candidate updates, state resetting, and joint acceptance. Dashed lines indicate on-demand calls to the shared mathematical tools.
    }
    \label{fig:biverify_framework}
\end{figure*}

BiVerify performs verification after the model has generated a complete initial solution. Given a mathematical problem \(q\) and initial solution \(y_0\), BiVerify extracts an answer and maintains \(C_t=(q,y_t,a_t)\) at verification event \(t\), where \(y_t\) and \(a_t\) are the current complete solution and final answer. As Figure~\ref{fig:biverify_framework} shows, backward and forward verification operate on the same candidate, while a scheduler coordinates direction switching, candidate updates, and verification-flow refinement according to verification outcomes. Both directions share mathematical tools, and acceptance requires the same candidate to pass both. The complete pseudocode for BiVerify is provided in Supplementary Algorithm~\ref{alg:biverify}.

\subsection{Backward Verification}

Backward verification starts from \(a_t\), checks it against the original conditions in \(q\), and retains \(y_t\) as context. It asks whether \(a_t\) satisfies the necessary conditions.

It first constructs a backward verification flow:
\[
V_t^{B}=\mathcal A_{\mathrm{flow}}^{B}(q,a_t).
\]
Here, \(\mathcal A_{\mathrm{flow}}^{B}\) is the backward verification-flow agent, and \(V_t^{B}\) specifies how to check \(a_t\) against the necessary constraints in \(q\). Depending on the problem, \(V_t^{B}\) may substitute \(a_t\); check equations, identities, domains, ranges, or feasibility; and test counting, extrema, or other answer-related conditions. These examples are not exhaustive.

The backward flow and adjudication agents may invoke the shared mathematical tool agent \(\mathcal A_{\mathrm{tool}}\), which dispatches requests to
\[
\mathcal T=
\{
T_{\mathrm{sym}},
T_{\mathrm{eq}},
T_{\mathrm{enum}},
T_{\mathrm{con}},
T_{\mathrm{arith}},
\ldots
\},
\]
where \(T_{\mathrm{sym}}\), \(T_{\mathrm{eq}}\), \(T_{\mathrm{enum}}\), \(T_{\mathrm{con}}\), and \(T_{\mathrm{arith}}\) denote symbolic simplification, equation solving, enumeration and counting, constraint checking, and exact computation. Backward verification mainly uses \(T_{\mathrm{eq}}\), \(T_{\mathrm{con}}\), and \(T_{\mathrm{arith}}\) for answer-related conditions. SymPy and Z3 implement symbolic computation and constraint solving, respectively~\citepboxed{meurer2017sympy,demoura2008z3}. Tool outputs provide evidence, not verdicts; subsequent agent arguments omit \(\mathcal A_{\mathrm{tool}}\) and \(\mathcal T\) for brevity.

Given candidate \(C_t\) and flow \(V_t^{B}\), backward adjudication returns
\[
(B_t,E_t^{B})
=
\mathcal A_{\mathrm{ver}}^{B}
(C_t,V_t^{B}),
\]
where \(\mathcal A_{\mathrm{ver}}^{B}\) is the backward adjudication agent, \(B_t\) the verdict, and \(E_t^{B}\) the verification trace, tool outputs, and failure evidence. The verdict is
\[
B_t=
\begin{cases}
\mathrm{Pass}, & \text{all required checks pass},\\
\mathrm{Fail}, & \text{a required check fails}.
\end{cases}
\]

On failure, \(E_t^{B}\) records the conflict between the answer and original constraints---such as a violated equation, constraint, range, or numerical relation---with the corresponding tool outputs. The scheduler converts \(E_t^{B}\) into feedback for revising \(C_t\) or \(V_t^{B}\).

\subsection{Forward Verification}

Forward verification takes \(q\), \(y_t\), and \(a_t\) as input and checks whether the answer follows from correct, complete reasoning over the problem conditions and current solution. Rather than testing constraints around the answer, it examines how the answer follows from those conditions and intermediate reasoning.

It first constructs ordered key claims and a corresponding verification flow:
\[
(K_t,V_t^{F})
=
\mathcal A_{\mathrm{flow}}^{F}(q,y_t,a_t),
\]
where
\[
K_t=
(c_{t,1},c_{t,2},\ldots,c_{t,n_t})
\]
Here, \(\mathcal A_{\mathrm{flow}}^{F}\) is the forward verification-flow agent; \(K_t\) contains \(n_t\) claims in their original order; and \(V_t^{F}\) specifies their verification order, objectives, and on-demand tool tasks.

The forward flow and adjudication agents may likewise invoke \(\mathcal A_{\mathrm{tool}}\), mainly for symbolic simplification, equation solving, enumeration and counting, and exact computation when checking intermediate claims and completeness; tool outputs accompany the evidence.

Key claims may cover algebraic or geometric derivations, numerical or symbolic calculations, condition transformations, domain or constraint derivations, intermediate conclusions, and the final answer. Requirements for all solutions, complete counts, exhaustive enumerations, complete sets, or global extrema are likewise included in \(K_t\) rather than handled by separate final-answer post-processing.

Forward verification checks claims in order. Given \(q\) and prior verified claims, the idealized objective for \(c_{t,i}\) is
\[
q\land c_{t,1}\land\cdots\land c_{t,i-1}
\models
c_{t,i}.
\]
Here, \(\models\) denotes semantic entailment: a claim passes if every case satisfying the problem conditions and prior verified claims also satisfies it.

If the first \(k_t-1\) claims pass and claim \(k_t\) fails, then
\[
c_{t,k_t}
\]
is the first failed claim. Because later reasoning may depend on it, subsequent claims are not treated as reliably supported; this claim and its evidence guide correction. A failure may instead indicate an inadequate verification flow, so the scheduler distinguishes candidate from flow errors.

Forward adjudication returns
\[
(F_t,E_t^{F},k_t)
=
\mathcal A_{\mathrm{ver}}^{F}
(C_t,K_t,V_t^{F}),
\]
where \(\mathcal A_{\mathrm{ver}}^{F}\) is the forward adjudication agent, \(F_t\) is its verdict, \(E_t^{F}\) contains the verification trace and tool evidence, and \(k_t\) is the first failed-claim position, omitted on a pass:
\[
F_t=
\begin{cases}
\mathrm{Pass}, & \text{all claims pass; objective complete},\\
\mathrm{Fail}, & \text{a claim fails or objective incomplete}.
\end{cases}
\]
On failure, the first failed claim \(c_{t,k_t}\) and evidence \(E_t^{F}\) guide revision of the candidate \(C_t\) or forward flow \(V_t^{F}\); the scheduler handles candidate updates, state resets, and direction selection.

\subsection{Bidirectional Verification Scheduler}

The bidirectional scheduler \(\mathcal S\) closes the two directions over the same candidate. Each adjudication agent returns a directional verdict, shared mathematical tools provide only evidence, and \(\mathcal S\) determines acceptance.

BiVerify extracts the initial answer from \(q\) and \(y_0\):
\[
a_0=\operatorname{Extract}(y_0).
\]
For the current candidate \(C_t\), the scheduling state is
\[
\Sigma_t=(C_t,B_t,F_t,d_t),
\]
where \(B_t\) and \(F_t\) are its backward and forward verification states:
\[
B_t,F_t\in
\{\mathrm{Unknown},\mathrm{Pass},\mathrm{Fail}\}.
\]
\(\mathrm{Unknown}\) means the direction is unverified or a candidate change invalidated its prior state; \(\mathrm{Pass}\) and \(\mathrm{Fail}\) record the outcome. States are candidate-specific and cannot transfer to a new candidate.

The current direction is
\[
d_t\in\{\mathrm{B},\mathrm{F}\},
\]
where \(\mathrm B\) and \(\mathrm F\) denote backward and forward verification. Acceptance requires both directions to pass:
\[
\operatorname{Accept}(C_t)
\iff
B_t=\mathrm{Pass}
\land
F_t=\mathrm{Pass}.
\]
Both pass states must belong to the same \(C_t\); their event indices may differ, but states combine only when bound to one candidate. If
\[
C_i\neq C_j,
\qquad
B_i=\mathrm{Pass},
\qquad
F_j=\mathrm{Pass},
\]
then neither candidate has completed bidirectional verification, necessitating candidate updates and state resets.

Initially,
\[
\begin{aligned}
C_0&=(q,y_0,a_0),
& B_0&=F_0=\mathrm{Unknown},\\
d_0&=\mathrm{B},
\end{aligned}
\]
so verification begins in the backward direction.

Let \(s_t\in\{\mathrm{Pass},\mathrm{Fail}\}\) be the current verdict, which the scheduler records in \(B_t\) or \(F_t\) according to \(d_t\). On failure, it invokes the verification-state and candidate-update agent with the candidate, evidence, and direction:
\[
\widehat C_t
=
\mathcal A_{\mathrm{update}}
(C_t,E_t^{d_t},d_t),
\]
\(\mathcal A_{\mathrm{update}}\) converts the failure evidence \(E_t^{d_t}\) into feedback, revises the solution \(y_t\), and re-extracts its answer to produce a proposal \(\widehat C_t\). Feedback may include a failed constraint or the first failed forward claim \(c_{t,k_t}\), tool outputs, and a correction direction. The proposal \(\widehat C_t\) is returned to the scheduler \(\mathcal S\), which adopts it only if valid: \(\widehat C_t=C_t\) denotes no adoption, whereas \(\widehat C_t\neq C_t\) denotes adoption. The proposal is unused when \(s_t=\mathrm{Pass}\).

The scheduler follows
\[
\begin{aligned}
&\mathcal S(\Sigma_t,s_t,\widehat C_t)\\[-1mm]
&\quad=
\begin{cases}
\mathrm{Accept},
& B_t=F_t=\mathrm{Pass},\\[1mm]
\mathrm{Switch},
& s_t=\mathrm{Pass}\land\neg\operatorname{Accept}(C_t),\\[1mm]
\mathrm{Update},
& s_t=\mathrm{Fail}\land\widehat C_t\neq C_t,\\[1mm]
\mathrm{Refine},
& s_t=\mathrm{Fail}\land\widehat C_t=C_t.
\end{cases}
\end{aligned}
\]
Here, \(\mathrm{Accept}\) accepts the candidate and terminates; \(\mathrm{Switch}\) moves to the opposite direction \(\operatorname{Other}(d_t)\); \(\mathrm{Update}\) adopts the new candidate, resets both states, and re-verifies the current direction; and \(\mathrm{Refine}\) revises its verification flow before retrying it. Acceptance is terminal, not a value of \(d_t\).

Under \(\mathrm{Update}\), the scheduler adopts the revision:
\[
C_{t+1}
=
(q,y_{t+1},a_{t+1})
=
\widehat C_t.
\]
Because prior states belong to the previous candidate, both reset:
\[
B_{t+1}
=
F_{t+1}
=
\mathrm{Unknown}.
\]
A flow bound to the previous candidate must also be regenerated or revised, while the failed direction is retained:
\[
d_{t+1}=d_t,
\]
to confirm that the revision resolves the problem.

Under \(\mathrm{Refine}\), no new candidate is adopted:
\[
C_{t+1}=C_t.
\]
Failure may instead reflect omitted checks, a misspecified objective, insufficient tool tasks, or a flow--candidate mismatch. Accordingly, \(\mathcal A_{\mathrm{update}}\) uses \(C_t\) and \(E_t^{d_t}\) to revise and retry that flow:
\[
d_{t+1}=d_t.
\]
This separates solution from flow revision and prevents premature switching.

If the budget ends before \(B_t=F_t=\mathrm{Pass}\), the implementation may return the latest \(C_t\) as a budget-limited output.

\begin{table*}[t]
\centering
\small
\setlength{\tabcolsep}{2.8pt}
\renewcommand{\arraystretch}{1.08}
\begin{tabular}{lcccccccc}
\toprule
Method & DS-Flash & DS-Flash-Think & DS-Pro & GPT-Mini & GPT-Mini-Think & Gemini-Lite & Gemini-Lite-Think & Mean \\
\midrule
Initial Score & 44.1 & 88.8 & 38.8 & 22.9 & 50.6 & 36.5 & 37.6 & 45.6 \\
Reflex & \(45.3_{\scoreup{+1.2}}\) & \(84.1_{\scoredown{-4.7}}\) & \(44.1_{\scoreup{+5.3}}\) & \(32.4_{\scoreup{+9.5}}\) & \(60.6_{\scoreup{+10.0}}\) & \(38.2_{\scoreup{+1.7}}\) & \(41.8_{\scoreup{+4.2}}\) & \(49.5_{\scoreup{+3.9}}\) \\
Self-Refine & \(50.6_{\scoreup{+6.5}}\) & \(83.5_{\scoredown{-5.3}}\) & \(53.5_{\scoreup{+14.7}}\) & \(39.4_{\scoreup{+16.5}}\) & \(60.6_{\scoreup{+10.0}}\) & \(36.5_{\scorezero{+0}}\) & \(41.2_{\scoreup{+3.6}}\) & \(52.2_{\scoreup{+6.6}}\) \\
Self-Reflection & \(45.3_{\scoreup{+1.2}}\) & \(86.5_{\scoredown{-2.3}}\) & \(44.7_{\scoreup{+5.9}}\) & \(34.7_{\scoreup{+11.8}}\) & \(65.3_{\scoreup{+14.7}}\) & \(37.1_{\scoreup{+0.6}}\) & \(40.6_{\scoreup{+3.0}}\) & \(50.6_{\scoreup{+5.0}}\) \\
CheckList & \(44.7_{\scoreup{+0.6}}\) & \(81.8_{\scoredown{-7.0}}\) & \(45.3_{\scoreup{+6.5}}\) & \(33.5_{\scoreup{+10.6}}\) & \(60.6_{\scoreup{+10.0}}\) & \(37.1_{\scoreup{+0.6}}\) & \(38.8_{\scoreup{+1.2}}\) & \(48.8_{\scoreup{+3.2}}\) \\
CoT-Tool & \(45.9_{\scoreup{+1.8}}\) & \(77.6_{\scoredown{-11.2}}\) & \(31.2_{\scoredown{-7.6}}\) & \(30.6_{\scoreup{+7.7}}\) & \(54.7_{\scoreup{+4.1}}\) & \(27.6_{\scoredown{-8.9}}\) & \(37.6_{\scorezero{+0}}\) & \(43.6_{\scoredown{-2.0}}\) \\
ProgCo & \(51.2_{\scoreup{+7.1}}\) & \(87.6_{\scoredown{-1.2}}\) & \(44.7_{\scoreup{+5.9}}\) & \(24.7_{\scoreup{+1.8}}\) & \(57.6_{\scoreup{+7.0}}\) & \(37.6_{\scoreup{+1.1}}\) & \(40.0_{\scoreup{+2.4}}\) & \(49.1_{\scoreup{+3.5}}\) \\
ProgCo-Py & \(52.9_{\scoreup{+8.8}}\) & \(88.2_{\scoredown{-0.6}}\) & \(49.4_{\scoreup{+10.6}}\) & \(24.7_{\scoreup{+1.8}}\) & \(59.4_{\scoreup{+8.8}}\) & \(35.9_{\scoredown{-0.6}}\) & \(41.8_{\scoreup{+4.2}}\) & \(50.3_{\scoreup{+4.7}}\) \\
\midrule
Backward-only & \(61.2_{\scoreup{+17.1}}\) & \(\mathbf{89.4}_{\scoreup{+0.6}}\) & \(57.1_{\scoreup{+18.3}}\) & \(29.4_{\scoreup{+6.5}}\) & \(61.2_{\scoreup{+10.6}}\) & \(40.0_{\scoreup{+3.5}}\) & \(42.4_{\scoreup{+4.8}}\) & \(54.4_{\scoreup{+8.8}}\) \\
Forward-only & \(60.6_{\scoreup{+16.5}}\) & \(86.5_{\scoredown{-2.3}}\) & \(69.4_{\scoreup{+30.6}}\) & \(\mathbf{70.6}_{\scoreup{+47.7}}\) & \(63.5_{\scoreup{+12.9}}\) & \(40.0_{\scoreup{+3.5}}\) & \(39.4_{\scoreup{+1.8}}\) & \(61.4_{\scoreup{+15.8}}\) \\
\textbf{BiVerify} & \(\mathbf{71.2}_{\scoreup{+27.1}}\) & \(88.8_{\scorezero{+0}}\) & \(\mathbf{75.9}_{\scoreup{+37.1}}\) & \(\mathbf{70.6}_{\scoreup{+47.7}}\) & \(\mathbf{66.5}_{\scoreup{+15.9}}\) & \(\mathbf{41.2}_{\scoreup{+4.7}}\) & \(\mathbf{45.9}_{\scoreup{+8.3}}\) & \(\mathbf{65.7}_{\scoreup{+20.1}}\) \\
\bottomrule
\end{tabular}
\caption{Average-accuracy overview of the main experiments. Each model column reports sample-weighted average accuracy over the five challenging benchmarks in the corresponding setting. Subscripts denote absolute changes from Initial Score, with green and red indicating nonnegative and negative changes, respectively. Mean is the average of Avg over the seven model settings. Bold denotes the highest or tied-highest value in each column.}
\label{tab:main_avg_results}
\end{table*}

 \section{Experiments}

We analyze BiVerify at three levels: overall benchmark and model performance; comparisons with ProgCo and ProgCo-Py by iteration budget, empirical difficulty, and correction behavior; and internal mechanisms through unidirectional components, event-level accuracy, directional roles, and correction stability. These analyses assess its effectiveness, sources of advantage, and operation.

\subsection{Experimental Setup}\label{sec:experiment_setup}

\noindent\textbf{Datasets.}
We evaluate five challenging benchmarks totaling 170 problems. AIME 2024~\citepboxed{huggingfaceh42025aime2024} and AIME 2025~\citepboxed{yentinglin2025aime} each contain 30 American Invitational Mathematics Examination problems testing multi-step reasoning and exact computation. BRUMO 2025 has 30 problems with complex conditions and multi-step reasoning; HMMT February 2025 has 30 spanning algebra, geometry, combinatorics, and number theory~\citepboxed{balunovic2025matharena}. AMO Bench has 50 Olympiad-level problems with diverse problem and answer formats~\citepboxed{an2025amobench}. We abbreviate them as A24, A25, B25, HMM, and AMO; cross-benchmark Avg values are sample-weighted.

\noindent\textbf{Base Models.}
We use seven DeepSeek, GPT, and Gemini settings: DeepSeek-Flash, DeepSeek-Flash-Think, and DeepSeek-Pro~\citepboxed{deepseekai2026v4}; GPT-5.4-Mini and GPT-5.4-Mini-Think~\citepboxed{openai2026gpt54systemcard}; and Gemini-2.5-Flash-Lite and Gemini-2.5-Flash-Lite-Think~\citepboxed{gemini2025technicalreport,geminiModelsDocs}. We abbreviate them as DS-Flash, DS-Flash-Think, DS-Pro, GPT-Mini, GPT-Mini-Think, Gemini-Lite, and Gemini-Lite-Think. They span standard and thinking modes, stronger variants, model families, and initial capabilities.

\noindent\textbf{Compared Methods.}
We compare three public-baseline categories. Natural-language correction includes Reflex, following ProgCo's vanilla-reflex setting~\citepboxed{song-etal-2025-progco}; Self-Refine with iterative self-feedback~\citepboxed{madaan2023selfrefine}; Self-Reflection with reflective feedback~\citepboxed{shinn2023reflexion}; and CheckList with checklist-guided correction~\citepboxed{cook2024ticking}. CoT-Tool uses multi-path chain-of-thought sampling and consistency selection~\citepboxed{abedi_cogitator_2025,wei2022chain,wang2022selfconsistency}. Verification-enhanced baselines are ProgCo, with program-driven verification and refinement, and ProgCo-Py, with Python-tool feedback~\citepboxed{song-etal-2025-progco}. We also introduce Backward-only and Forward-only, which use only the respective direction, while full BiVerify follows its scheduler. These component ablations isolate each direction under identical base-model, initial-response, and iteration-budget settings. Figures and tables use Backward-only and Forward-only; execution directions use Backward and Forward.

For each base-model--problem pair, we pre-generate an initial solution and extract its answer. All methods except CoT-Tool share this initial solution and extracted answer; CoT-Tool samples three reasoning paths from the original problem. Within each setting, methods use the same model identifier and reasoning mode, with temperature \(0\) when applicable. All iterative methods use \(\texttt{max\_turn}=3\). BiVerify starts backward, and log indices \(t=0,\ldots,3\) represent at most four actual verification events. Supplementary Tables~\ref{tab:app_fairness_settings} and~\ref{tab:app_model_configurations} report shared configurations and model identifiers.

\noindent\textbf{Evaluation Metrics.}
Our primary metric is final-answer accuracy. Empty outputs, extraction failures, and reference mismatches count as incorrect under the original benchmark-size denominator. For each model, Avg is sample-weighted across benchmarks, and Mean averages Avg across all seven settings. Let \(C\) and \(W\) denote correct and incorrect initial or current answers. Then \(W\rightarrow C\) is correction, \(C\rightarrow W\) is damage to an initially correct answer, and \(N_{W\rightarrow C}-N_{C\rightarrow W}\) is net correction gain. Event analysis follows log order; after early termination, the terminal answer is carried forward, retaining 1,190 model--problem instances at every stage.

\subsection{Main Results}\label{sec:main_results}
Table~\ref{tab:main_avg_results} reports Avg for all seven settings and their Mean; subscripts give changes from the corresponding Initial Score.
BiVerify averages \(65.7\%\), exceeding the \(45.6\%\) Initial Score and the strongest evaluated public baseline at \(52.2\%\) by \(20.1\) and \(13.5\) points. It also exceeds backward-only and forward-only verification by \(11.3\) and \(4.3\) points, showing that closing the loop between answer-centered backward and process-centered forward verification outperforms either alone.

BiVerify outperforms the strongest evaluated public baseline in all seven settings. Its largest gain is on GPT: \(70.6\%\), or \(31.2\) points above the strongest evaluated baseline at \(39.4\%\). Gains across DeepSeek and Gemini span model families and initial capabilities. The smaller DS-Flash-Think gain reflects its high \(88.8\%\) Initial Score and limited headroom.

Backward-only and forward-only lead in different settings, suggesting partly distinct error coverage. BiVerify matches or exceeds the stronger variant in most settings and has a higher Mean, supporting same-candidate closed-loop re-verification.

BiVerify improves all five benchmarks, most on HMM and AIME 2025 by about \(26.7\) and \(24.8\) points; thus, no single model or benchmark drives its advantage. Supplementary Tables~\ref{tab:appendix_detailed_deepseek},~\ref{tab:appendix_detailed_gpt}, and~\ref{tab:appendix_detailed_gemini} give complete seven-setting, five-benchmark results and further analyses.

\begin{figure}[t]
\centering
\includegraphics[width=0.90\linewidth]{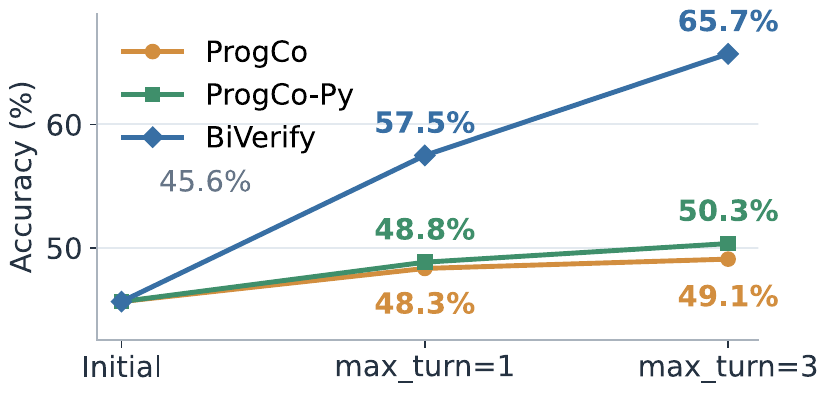}
\caption{Accuracy comparison of tool-augmented methods under different iteration budgets. All three methods use the same initial responses. The horizontal axis shows Initial, \(\texttt{max\_turn}=1\), and \(\texttt{max\_turn}=3\); the vertical axis reports average accuracy over the seven base-model settings. Equal \(\texttt{max\_turn}\) values indicate only the same outer iteration limit, not identical internal model or tool calls across methods.}
\label{fig:baseline_budget_accuracy}
\end{figure}

\subsection{Further Comparison with Tool-Augmented Baselines}\label{sec:tool_augmented_baseline_analysis}
We compare BiVerify with the related tool-augmented methods ProgCo and ProgCo-Py by iteration budget, correction behavior, and empirical difficulty, using shared initial responses and identical model--problem instances.

\noindent\textbf{Iteration Budget and Accuracy.} Figure~\ref{fig:baseline_budget_accuracy} shows when gains emerge by comparing ProgCo, ProgCo-Py, and BiVerify from the same Initial point across iteration limits on 1,190 instances. BiVerify exceeds the stronger baseline by \(8.7\) points at \(\texttt{max\_turn}=1\) and \(15.4\) at \(\texttt{max\_turn}=3\).

BiVerify's advantage appears early and grows across iterations. Equal outer limits compare accuracy, not runtime, tokens, or tool-call efficiency.

\begin{figure}[t]
\centering
\includegraphics[width=0.95\linewidth]{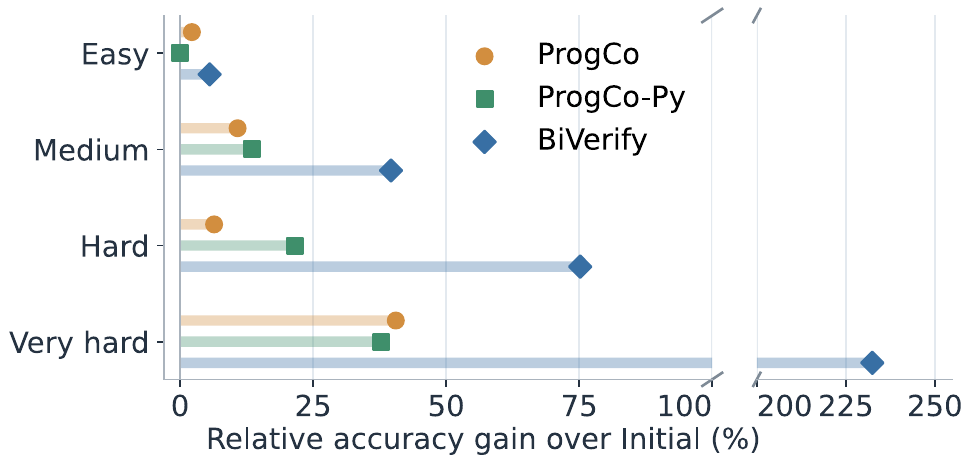}
\caption{Accuracy gains over Initial by empirical difficulty. Difficulty is determined by the number of the seven base models that initially answer a problem correctly. The horizontal axis reports relative accuracy gain rather than absolute percentage-point change; to reduce horizontal space, the axis is broken between \(100\%\) and \(200\%\).}
\label{fig:difficulty_relative_gain}
\end{figure}

\noindent\textbf{Performance by Empirical Difficulty.} We define empirical difficulty by the number of seven base models initially solving each problem: Easy, Medium, Hard, and Very hard correspond to 6--7, 4--5, 2--3, and 0--1. Figure~\ref{fig:difficulty_relative_gain} compares error recovery across initial accuracies using relative gain, \((A_{\mathrm{final}}-A_{\mathrm{initial}})/A_{\mathrm{initial}}\times100\%\).

BiVerify's relative gain rises with difficulty, reaching \(232.4\%\) for Very hard versus \(40.5\%\) for the stronger programmatic-verification baseline, concentrating gains where initial failures are common. Because low initial accuracy amplifies relative changes, Supplementary Tables~\ref{tab:appendix_detailed_deepseek}--\ref{tab:appendix_detailed_gemini} give complete absolute accuracies.

We compare final correctness transitions under shared initial responses and fixed denominators, separating \(W\rightarrow C\) corrections from \(C\rightarrow W\) damage. BiVerify corrects \(40.3\%\) of initially incorrect answers, versus \(9.4\%\) for ProgCo and \(12.2\%\) for ProgCo-Py; their respective damage rates on initially correct answers are \(4.1\%\), \(3.7\%\), and \(4.2\%\). This gap shows that BiVerify's advantage comes from broader correction coverage rather than more aggressive rewriting. Supplementary Figure~\ref{fig:baseline_transition_flow} gives complete four-way flows, rates, and counts.

\subsection{Internal Mechanism Analysis}\label{sec:internal_mechanism_analysis}
We next examine BiVerify internally through unidirectional components, event-level accuracy, directional corrections, and final-correction stability.

\noindent\textbf{Unidirectional Verification and Bidirectional Combination.} BiVerify corrects 261 initially incorrect answers and damages 22 initially correct ones, a net gain of 239. It adds 51 final corrections over Forward-only with the same 22 damaged answers, showing that bidirectional verification expands correction coverage without more aggressive rewriting. The resulting \(20.1\)-point gain equals the Initial-to-final improvement, linking aggregate and sample-level results. Supplementary Table~\ref{tab:direction_variant_transition} gives the full comparison with Backward-only and Forward-only.

\begin{figure}[t]
\centering
\includegraphics[width=0.90\linewidth]{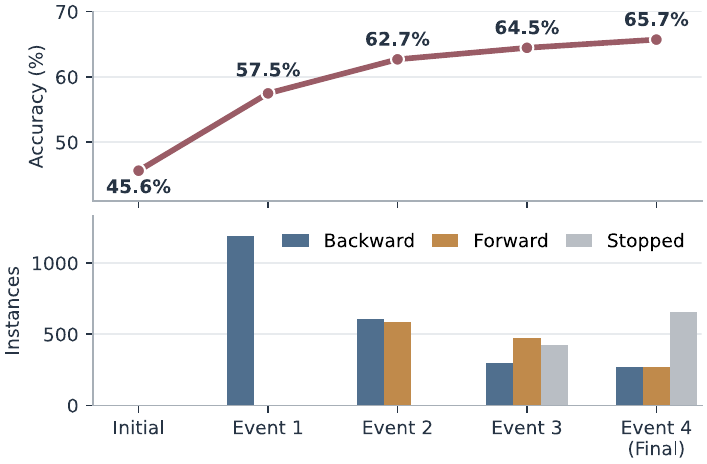}
\caption{Accuracy and scheduling over actual verification events. Initial precedes verification; Event \(k\) is each sample's \(k\)-th executed event. Event 1 is backward for all samples; later events follow scheduler feedback. The upper panel reports accuracy over 1,190 instances; the lower counts backward, forward, and terminated instances. Terminated answers are carried forward, and Event 4 is Final.}
\label{fig:event_trajectory}
\end{figure}

\noindent\textbf{Accuracy Evolution across Verification Events.} To examine how gains accumulate within bidirectional verification, Figure~\ref{fig:event_trajectory} reconstructs BiVerify's accuracy trajectory over the verification events that actually occur. Overall accuracy rises from \(45.6\%\) at Initial to \(65.7\%\) at the final state, with the first two verification events accounting for \(84.6\%\) of the total improvement. Most gains therefore emerge early; later events continue to correct unresolved samples, albeit with diminishing marginal gains. We analyze the contribution of each verification direction separately below.

\begin{figure}[!t]
\centering
\includegraphics[width=0.87\linewidth]{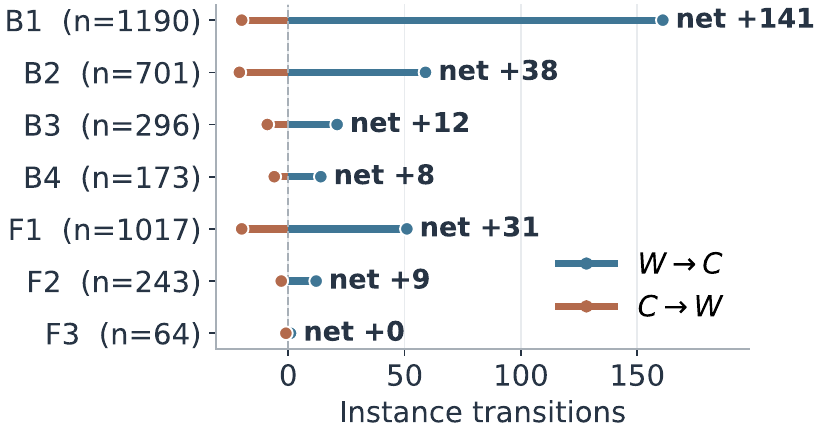}
\caption{Corrections and damage by within-sample directional occurrence in BiVerify. B\(k\) and F\(k\) denote the \(k\)-th backward and forward verification event within a sample, rather than a global event position. Blue bars extending right indicate \(W\rightarrow C\), orange bars extending left indicate \(C\rightarrow W\), and terminal Net is defined as \(N_{W\rightarrow C}-N_{C\rightarrow W}\).}
\label{fig:direction_contribution}
\end{figure}

Scheduling is not fixed: all samples begin with backward verification; feedback then triggers same-direction re-verification, a direction switch, or early termination. By Event 4, \(55.0\%\) of the instances have terminated. Different samples therefore follow verification paths of different lengths and directional compositions; the complete path distribution is shown in Supplementary Figure~\ref{fig:path_analysis}.

\noindent\textbf{Roles of Backward and Forward Verification.} Figure~\ref{fig:direction_contribution} reports corrections and damage for B\(k\) and F\(k\), the \(k\)-th within-sample occurrences of their respective directions; these are not independent Backward-only and Forward-only runs. Across their respective occurrences, backward and forward verification yield cumulative net gains of \(+199\) and \(+40\), respectively, showing asymmetric but complementary correction coverage.

Later re-verification events still yield positive gains, although marginal gains decline at later same-direction occurrences. Among the 261 final \(W\!\rightarrow\!C\) instances, 248 remain correct after first becoming correct, so stable repairs account for \(95.0\%\). Another 27 instances are temporarily corrected but ultimately return to incorrect; including these cases, 288 initially incorrect instances become correct at least once, of which 261 remain correct at the end, yielding a \(90.6\%\) final retention rate. In addition, 16 instances recover after intermediate damage. Thus, the overall gain comes primarily from corrections retained to the end, although later events can introduce limited fluctuations. Complete correction-stability and cross-model trajectory analyses are provided in Supplementary Figures~\ref{fig:trajectory_classes} and~\ref{fig:cross_model_trajectory}.

 \section{Conclusion}

We introduced BiVerify, a verification-feedback-driven framework for bidirectional mathematical verification. BiVerify applies answer-centered backward verification and process-centered forward verification to the same candidate and forms a closed loop through direction switching, state resetting, and re-verification, accepting only candidates that pass in both directions. Both directions use shared mathematical tools for operations such as symbolic computation and equation solving. Experiments on five challenging mathematical reasoning benchmarks and seven base-model settings show that BiVerify achieves higher overall accuracy than backward-only verification, forward-only verification, and the evaluated public baselines. Its main gains come from broader error-correction coverage without materially increasing damage to initially correct answers. In principle, BiVerify can be extended to tasks with both verifiable final outputs and inspectable intermediate processes. Future work will explore its application to scientific reasoning and code generation.

\bibliography{BiVerify_AAAI27}

\clearpage
\onecolumn
\raggedbottom
\graphicspath{{supplement_figures/}{Figures/BiVerify/}}
\appendix
\setcounter{secnumdepth}{2}
\begin{center}
{\LARGE\bfseries Supplementary Material}
\end{center}
\section{Experimental Settings and Detailed Results}
This supplementary material reports complete benchmark-level results for the main experiments at \(\texttt{max\_turn}=3\), organized by the DeepSeek, GPT, and Gemini model families. In the detailed accuracy tables, all values are accuracies (\%), and Avg is the sample-weighted average over the five benchmarks. Bold denotes the highest or tied-highest value among correction methods for each model and benchmark column; Initial Score is excluded from this marking.

\subsection{Complete Run Configuration and Reproducibility Details}\label{app:reproducibility}

\noindent\textbf{Fairness Controls.}
For each base-model--problem pair, all methods invoke the same official model identifier for that experimental setting and use the same standard-generation or thinking mode; when applicable, the temperature is fixed at \(0\). Except for CoT-Tool, which generates candidates from the original problem using its own multi-path mechanism, all iterative correction and verification methods receive the same pre-generated initial solution and candidate answer. No method gains an advantage by substituting a stronger base model. Table~\ref{tab:app_fairness_settings} summarizes the shared model configuration, input, and outer-budget protocol for each method category.

\begin{table}[!htbp]
\centering
\small
\setlength{\tabcolsep}{5pt}
\renewcommand{\arraystretch}{1.08}
\begin{tabularx}{0.96\textwidth}{X c c X l}
\toprule
Method & Base Model and Reasoning Mode & Temp. & Starting Input & Outer Budget \\
\midrule
Initial Score
& Same within setting & 0 & Shared initial generation & No post-processing \\
Reflex / Self-Refine / Self-Reflection / CheckList
& Same within setting & 0 & Same initial solution & \(\texttt{max\_turn}=3\) \\
ProgCo / ProgCo-Py
& Same within setting & 0 & Same initial solution & \(\texttt{max\_turn}=3\) \\
Backward-only / Forward-only / BiVerify
& Same within setting & 0 & Same initial solution & \(\texttt{max\_turn}=3\) \\
CoT-Tool
& Same within setting & 0 & Same original problem & \(\texttt{n\_samples}=3\) \\
\bottomrule
\end{tabularx}
\caption{Fairness controls across methods. Same within setting means that methods use the same official model identifier and reasoning mode for each base-model--problem pair; CoT-Tool follows its native multi-path protocol.}
\label{tab:app_fairness_settings}
\end{table}

\noindent\textbf{Base Models and Reasoning Modes.}
Table~\ref{tab:app_model_configurations} lists the actual model identifiers and reasoning configurations for the seven experimental settings. Each row is held constant across all methods in that setting. Names with the \texttt{-Think} suffix are experimental labels used to distinguish reasoning configurations; standard and thinking modes of the same model use the same base model or the corresponding thinking endpoint and are distinguished through API reasoning parameters.

\begin{table}[!htbp]
\centering
\small
\setlength{\tabcolsep}{5pt}
\renewcommand{\arraystretch}{1.05}
\begin{tabularx}{0.94\textwidth}{l l X}
\toprule
Paper Setting & Actual Model Identifier & Reasoning Configuration \\
\midrule
DS-Flash & \texttt{deepseek-v4-flash} & Thinking disabled \\
DS-Flash-Think & \texttt{deepseek-v4-flash} & Thinking enabled, \texttt{reasoning\_effort=high} \\
DS-Pro & \texttt{deepseek-v4-pro} & Thinking disabled \\
GPT-Mini & \texttt{gpt-5.4-mini} & Thinking disabled \\
GPT-Mini-Think & \texttt{gpt-5.4-mini} & Thinking enabled, \texttt{reasoning\_effort=low} \\
Gemini-Lite & \texttt{gemini-2.5-flash-lite} & Standard generation \\
Gemini-Lite-Think & \texttt{gemini-2.5-flash-lite-thinking} & Thinking enabled, \texttt{reasoning\_effort=low} \\
\bottomrule
\end{tabularx}
\caption{Base-model identifiers and reasoning modes. Model abbreviations are consistent with the results table.}
\label{tab:app_model_configurations}
\end{table}

\noindent\textbf{Method Budgets and Tool Use.}
All iterative correction and verification methods use an outer budget of \(\texttt{max\_turn}=3\), while CoT-Tool generates three reasoning paths and applies consistency-based selection according to its method definition. This control runs the compared methods with the same base model, reasoning mode, and predefined outer budget, but does not force identical internal model-call or tool-call structures because these steps are part of each method's definition. Both verification directions and their mathematical branches in BiVerify use the same base model corresponding to the current experiment; the shared Python mathematical tools provide execution results and do not replace the base model.

BiVerify is evaluated without an extensive hyperparameter sweep. We compare \(\texttt{max\_turn}\in\{1,3\}\) in the iteration-budget analysis and use \(\texttt{max\_turn}=3\) for the main comparisons, thereby imposing a common outer-iteration limit on BiVerify and the iterative baselines. Unspecified settings follow the corresponding method implementation or model API defaults.

Run variation is controlled by supplying the same pregenerated starting response and candidate answer to methods that operate from an existing solution within each base-model--problem pair. The official execution path contains no randomized local sampling, shuffling, or preprocessing, so a local random seed is not needed. We retain one official run for every model--method--problem combination and include the raw outputs underlying the reported metrics in the supplementary material.

\noindent\textbf{Unified Answer Evaluation.}
Final answers from all methods are evaluated using the same three-stage equivalence procedure. We first perform normalized exact-string matching. If an answer can be parsed numerically, we additionally apply floating-point comparison with a tolerance of \(10^{-3}\). When the first two stages are inconclusive, we use DeepSeek-Flash as an answer-equivalence evaluator with temperature \(0\). For set-valued, multi-solution, or sequential answers, evaluation also checks element completeness and any required ordering constraints. The five benchmarks use fixed denominators of 30, 30, 30, 30, and 50, respectively. Empty outputs, unextractable answers, and failures after exhausting retries remain in the denominator and are counted as incorrect rather than skipped.

\noindent\textbf{Code Availability.}
Code is available at \coderepolink{https://github.com/TicusFFF/BiVerify}.

\subsection{Detailed Tables by Model Family}
Tables~\ref{tab:appendix_detailed_deepseek},~\ref{tab:appendix_detailed_gpt}, and~\ref{tab:appendix_detailed_gemini} expand the text Table~\ref{tab:main_avg_results} by benchmark, reporting the accuracy of every method on A24, A25, B25, HMM, and AMO for the DeepSeek, GPT, and Gemini model families, respectively. The main text analyzes cross-benchmark coverage, while the supplementary material retains complete values for tracing the source of each overall Avg.
\begin{table}[!htbp]
\centering
\small
\setlength{\tabcolsep}{6.0pt}
\renewcommand{\arraystretch}{0.96}
\begin{tabular}{llrrrrrr}
\toprule
Model & Method & A24 & A25 & B25 & HMM & AMO & Avg \\
\midrule
\multirow{11}{*}{DS-Flash} & Initial Score & 63.3 & 43.3 & 86.7 & 33.3 & 14.0 & 44.1 \\
 & Reflex & 63.3 & 46.7 & 83.3 & 33.3 & 18.0 & 45.3 \\
 & Self-Refine & 66.7 & 60.0 & 86.7 & 50.0 & 14.0 & 50.6 \\
 & Self-Reflection & 63.3 & 46.7 & 86.7 & 33.3 & 16.0 & 45.3 \\
 & CheckList & 60.0 & 56.7 & 80.0 & 30.0 & 16.0 & 44.7 \\
 & CoT-Tool & 76.7 & 60.0 & 73.3 & 43.3 & 4.0 & 45.9 \\
 & ProgCo & 66.7 & 53.3 & \textbf{96.7} & 43.3 & 18.0 & 51.2 \\
 & ProgCo-Py & 66.7 & 60.0 & 90.0 & 43.3 & 24.0 & 52.9 \\
\cmidrule(lr){2-8}
 & Backward-only & 80.0 & 66.7 & 93.3 & 60.0 & 28.0 & 61.2 \\
 & Forward-only & 70.0 & 56.7 & 90.0 & 50.0 & \textbf{46.0} & 60.6 \\
 & \textbf{BiVerify} & \textbf{86.7} & \textbf{76.7} & 90.0 & \textbf{80.0} & 42.0 & \textbf{71.2} \\
\midrule
\multirow{11}{*}{DS-Flash-Think} & Initial Score & 100.0 & 100.0 & 100.0 & 100.0 & 62.0 & 88.8 \\
 & Reflex & 96.7 & \textbf{100.0} & 93.3 & 93.3 & 56.0 & 84.1 \\
 & Self-Refine & 93.3 & \textbf{100.0} & 90.0 & 96.7 & 56.0 & 83.5 \\
 & Self-Reflection & 96.7 & \textbf{100.0} & 93.3 & \textbf{100.0} & 60.0 & 86.5 \\
 & CheckList & 96.7 & \textbf{100.0} & 86.7 & 86.7 & 56.0 & 81.8 \\
 & CoT-Tool & 93.3 & 93.3 & 93.3 & 86.7 & 44.0 & 77.6 \\
 & ProgCo & \textbf{100.0} & \textbf{100.0} & \textbf{100.0} & \textbf{100.0} & 58.0 & 87.6 \\
 & ProgCo-Py & \textbf{100.0} & \textbf{100.0} & \textbf{100.0} & \textbf{100.0} & 60.0 & 88.2 \\
\cmidrule(lr){2-8}
 & Backward-only & \textbf{100.0} & \textbf{100.0} & \textbf{100.0} & \textbf{100.0} & \textbf{64.0} & \textbf{89.4} \\
 & Forward-only & 86.7 & 96.7 & \textbf{100.0} & \textbf{100.0} & \textbf{64.0} & 86.5 \\
 & \textbf{BiVerify} & 96.7 & \textbf{100.0} & \textbf{100.0} & \textbf{100.0} & \textbf{64.0} & 88.8 \\
\midrule
\multirow{11}{*}{DS-Pro} & Initial Score & 63.3 & 53.3 & 60.0 & 26.7 & 10.0 & 38.8 \\
 & Reflex & 60.0 & 63.3 & 70.0 & 40.0 & 10.0 & 44.1 \\
 & Self-Refine & 66.7 & 66.7 & 86.7 & 56.7 & 16.0 & 53.5 \\
 & Self-Reflection & 66.7 & 60.0 & 73.3 & 40.0 & 8.0 & 44.7 \\
 & CheckList & 66.7 & 53.3 & 70.0 & 40.0 & 16.0 & 45.3 \\
 & CoT-Tool & 56.7 & 43.3 & 43.3 & 23.3 & 6.0 & 31.2 \\
 & ProgCo & 66.7 & 60.0 & 66.7 & 40.0 & 12.0 & 44.7 \\
 & ProgCo-Py & 80.0 & 60.0 & 73.3 & 43.3 & 14.0 & 49.4 \\
\cmidrule(lr){2-8}
 & Backward-only & 76.7 & 66.7 & \textbf{93.3} & 53.3 & 20.0 & 57.1 \\
 & Forward-only & \textbf{90.0} & 80.0 & 86.7 & \textbf{80.0} & 34.0 & 69.4 \\
 & \textbf{BiVerify} & 83.3 & \textbf{90.0} & \textbf{93.3} & \textbf{80.0} & \textbf{50.0} & \textbf{75.9} \\
\bottomrule
\end{tabular}
\caption{Detailed accuracy of the three DeepSeek settings on the five benchmarks. The full model names are DeepSeek-Flash, DeepSeek-Flash-Think, and DeepSeek-Pro.}
\label{tab:appendix_detailed_deepseek}
\end{table}
\begin{table}[!htbp]
\centering
\small
\setlength{\tabcolsep}{6.0pt}
\renewcommand{\arraystretch}{0.96}
\begin{tabular}{llrrrrrr}
\toprule
Model & Method & A24 & A25 & B25 & HMM & AMO & Avg \\
\midrule
\multirow{11}{*}{GPT-Mini} & Initial Score & 40.0 & 26.7 & 43.3 & 13.3 & 4.0 & 22.9 \\
 & Reflex & 46.7 & 36.7 & 46.7 & 30.0 & 14.0 & 32.4 \\
 & Self-Refine & 60.0 & 46.7 & 60.0 & 36.7 & 12.0 & 39.4 \\
 & Self-Reflection & 46.7 & 40.0 & 63.3 & 23.3 & 14.0 & 34.7 \\
 & CheckList & 50.0 & 43.3 & 53.3 & 20.0 & 14.0 & 33.5 \\
 & CoT-Tool & 50.0 & 43.3 & 43.3 & 23.3 & 8.0 & 30.6 \\
 & ProgCo & 40.0 & 30.0 & 43.3 & 16.7 & 6.0 & 24.7 \\
 & ProgCo-Py & 46.7 & 26.7 & 43.3 & 13.3 & 6.0 & 24.7 \\
\cmidrule(lr){2-8}
 & Backward-only & 53.3 & 40.0 & 50.0 & 20.0 & 2.0 & 29.4 \\
 & Forward-only & \textbf{86.7} & 73.3 & 76.7 & \textbf{83.3} & \textbf{48.0} & \textbf{70.6} \\
 & \textbf{BiVerify} & \textbf{86.7} & \textbf{90.0} & \textbf{90.0} & 66.7 & 40.0 & \textbf{70.6} \\
\midrule
\multirow{11}{*}{GPT-Mini-Think} & Initial Score & 73.3 & 63.3 & 70.0 & 50.0 & 18.0 & 50.6 \\
 & Reflex & 83.3 & \textbf{83.3} & 70.0 & 66.7 & 24.0 & 60.6 \\
 & Self-Refine & 80.0 & \textbf{83.3} & 66.7 & 66.7 & 28.0 & 60.6 \\
 & Self-Reflection & 83.3 & \textbf{83.3} & \textbf{86.7} & \textbf{76.7} & 24.0 & 65.3 \\
 & CheckList & 83.3 & 80.0 & 73.3 & 66.7 & 24.0 & 60.6 \\
 & CoT-Tool & 80.0 & 80.0 & 70.0 & 56.7 & 14.0 & 54.7 \\
 & ProgCo & 73.3 & \textbf{83.3} & 76.7 & 53.3 & 24.0 & 57.6 \\
 & ProgCo-Py & 76.7 & 73.3 & 80.0 & 60.0 & 28.0 & 59.4 \\
\cmidrule(lr){2-8}
 & Backward-only & 86.7 & \textbf{83.3} & 83.3 & 60.0 & 20.0 & 61.2 \\
 & Forward-only & 76.7 & \textbf{83.3} & 83.3 & 73.3 & 26.0 & 63.5 \\
 & \textbf{BiVerify} & \textbf{90.0} & \textbf{83.3} & 80.0 & 66.7 & \textbf{34.0} & \textbf{66.5} \\
\bottomrule
\end{tabular}
\caption{Detailed accuracy of the two GPT settings on the five benchmarks. The full model names are GPT-5.4-Mini and GPT-5.4-Mini-Think.}
\label{tab:appendix_detailed_gpt}
\end{table}
\begin{table}[!htbp]
\centering
\small
\setlength{\tabcolsep}{6.0pt}
\renewcommand{\arraystretch}{0.96}
\begin{tabular}{llrrrrrr}
\toprule
Model & Method & A24 & A25 & B25 & HMM & AMO & Avg \\
\midrule
\multirow{11}{*}{Gemini-Lite} & Initial Score & 56.7 & 50.0 & 53.3 & 30.0 & 10.0 & 36.5 \\
 & Reflex & 56.7 & \textbf{60.0} & 53.3 & 30.0 & \textbf{10.0} & 38.2 \\
 & Self-Refine & 56.7 & 50.0 & 53.3 & 30.0 & \textbf{10.0} & 36.5 \\
 & Self-Reflection & 56.7 & 53.3 & 53.3 & 33.3 & 8.0 & 37.1 \\
 & CheckList & 60.0 & 50.0 & 56.7 & 30.0 & 8.0 & 37.1 \\
 & CoT-Tool & 53.3 & 33.3 & 46.7 & 20.0 & 2.0 & 27.6 \\
 & ProgCo & 63.3 & 50.0 & \textbf{63.3} & 33.3 & 2.0 & 37.6 \\
 & ProgCo-Py & 53.3 & 46.7 & 56.7 & 33.3 & 8.0 & 35.9 \\
\cmidrule(lr){2-8}
 & Backward-only & \textbf{66.7} & 53.3 & 60.0 & 33.3 & 8.0 & 40.0 \\
 & Forward-only & \textbf{66.7} & 50.0 & \textbf{63.3} & 33.3 & 8.0 & 40.0 \\
 & \textbf{BiVerify} & 63.3 & 56.7 & \textbf{63.3} & \textbf{36.7} & 8.0 & \textbf{41.2} \\
\midrule
\multirow{11}{*}{Gemini-Lite-Think} & Initial Score & 56.7 & 36.7 & 66.7 & 33.3 & 12.0 & 37.6 \\
 & Reflex & 70.0 & 50.0 & 70.0 & 36.7 & 6.0 & 41.8 \\
 & Self-Refine & 63.3 & 50.0 & 70.0 & 36.7 & 8.0 & 41.2 \\
 & Self-Reflection & 60.0 & 53.3 & \textbf{73.3} & 33.3 & 6.0 & 40.6 \\
 & CheckList & 63.3 & 46.7 & 60.0 & 40.0 & 6.0 & 38.8 \\
 & CoT-Tool & 66.7 & \textbf{56.7} & 50.0 & 33.3 & 4.0 & 37.6 \\
 & ProgCo & 60.0 & 46.7 & 63.3 & \textbf{43.3} & 8.0 & 40.0 \\
 & ProgCo-Py & 66.7 & 46.7 & 70.0 & 40.0 & 8.0 & 41.8 \\
\cmidrule(lr){2-8}
 & Backward-only & 63.3 & 53.3 & \textbf{73.3} & 33.3 & 10.0 & 42.4 \\
 & Forward-only & 56.7 & 40.0 & 70.0 & 40.0 & 10.0 & 39.4 \\
 & \textbf{BiVerify} & \textbf{73.3} & 50.0 & \textbf{73.3} & \textbf{43.3} & \textbf{12.0} & \textbf{45.9} \\
\bottomrule
\end{tabular}
\caption{Detailed accuracy of the two Gemini settings on the five benchmarks. The full model names are Gemini-2.5-Flash-Lite and Gemini-2.5-Flash-Lite-Think.}
\label{tab:appendix_detailed_gemini}
\end{table}

The detailed tables further reveal benchmark-level performance across model families. Within the DeepSeek family, BiVerify yields particularly large gains under standard generation; for example, on HMM and AMO with DeepSeek-Pro, it improves over Initial Score by \(53.3\) and \(40.0\) percentage points, respectively. The thinking mode begins with higher initial accuracy and therefore leaves less room for overall improvement. The largest gains occur in the GPT family: under standard GPT generation, BiVerify improves by \(63.3\) and \(53.4\) percentage points on AIME 2025 and HMM, respectively. BiVerify also achieves the highest overall accuracy in both Gemini settings. These results show that its benefits span multiple model families and benchmarks, although the magnitude depends on the base model's initial capability and benchmark characteristics.

\subsection{Statistical Significance}\label{app:statistical_significance}

We conduct a paired statistical analysis using the results from seven base-model settings on the same 170 problems. BiVerify improves average accuracy over Self-Refine, the strongest evaluated public baseline overall, by 13.53 percentage points. Using 200,000 BCa bootstrap resamples at the problem level, we obtain a 95\% confidence interval of [11.01, 16.30] percentage points for this improvement. The paired test yields \(p<0.001\). These results support BiVerify's overall accuracy improvement under the evaluated settings.

\subsection{Computational Cost}\label{app:computational_cost}

We additionally evaluate computational cost on 170 problems using DeepSeek-Flash in non-thinking mode. The reported costs cover verification and correction after the shared initial solution, excluding initial-solution generation and external final-answer evaluation. Table~\ref{tab:app_computational_cost} reports per-problem means.

\begin{table}[!htbp]
\centering
\small
\setlength{\tabcolsep}{6pt}
\renewcommand{\arraystretch}{1.05}
\begin{tabular}{lrr}
\toprule
Metric & ProgCo-Py & BiVerify \\
\midrule
LLM calls & 21.96 & 27.82 \\
Total tokens (millions) & 1.078 & 0.534 \\
Python executions & 11.67 & 8.44 \\
Task completion time (min) & 20.2 & 7.2 \\
\bottomrule
\end{tabular}
\caption{Per-problem computational cost on 170 problems using DeepSeek-Flash in non-thinking mode.}
\label{tab:app_computational_cost}
\end{table}

Under the evaluated configurations, BiVerify uses more LLM calls but fewer tokens and Python executions, with a shorter mean task completion time. Thus, a higher call count does not necessarily imply greater token consumption. We control costs through tool-selection guidance, budget reminders, and reuse of successful results. In this experiment, 327 cache hits avoid repeated Python execution. Task completion time is measured from each problem's first model or tool request to the final returned result, including waiting time.

\section{Supplementary Process Analysis}
This section supplements the text analyses of baseline correction behavior and the bidirectional verification process. Unidirectional component transitions compare final correction and damage with the complete framework; baseline state flows compare error correction and damage across methods; multi-event trajectory classes assess whether corrections persist to the final state; actual scheduling paths explain why samples experience sequences of different lengths and directional compositions; and cross-model event trajectories examine whether aggregate event-level gains are consistent across base models.

\subsection{Unidirectional Component Transitions}
\noindent
\begin{minipage}[t]{0.485\textwidth}
\vspace{0pt}
Table~\ref{tab:direction_variant_transition} reports final correctness transitions for backward-only verification, forward-only verification, and BiVerify on the same 1,190 model--problem instances.

BiVerify adds 51 final error corrections over forward-only verification without increasing final damage to initially correct answers.
\end{minipage}\hfill
\begin{minipage}[t]{0.485\textwidth}
\vspace{0pt}
\centering
\small
\setlength{\tabcolsep}{3.5pt}
\begin{tabular}{lrrrrr}
\toprule
Method & \(C\!\rightarrow\!C\) & \(C\!\rightarrow\!W\) & \(W\!\rightarrow\!C\) & \(W\!\rightarrow\!W\) & Net \\
\midrule
Backward-only & 528 & 15 & 119 & 528 & 104 \\
Forward-only & 521 & 22 & 210 & 437 & 188 \\
\textbf{BiVerify} & \textbf{521} & 22 & \textbf{261} & \textbf{386} & \textbf{239} \\
\bottomrule
\end{tabular}
\captionof{table}{Final correctness transitions for backward verification, forward verification, and BiVerify. Net is defined as \(N_{W\rightarrow C}-N_{C\rightarrow W}\).}
\label{tab:direction_variant_transition}
\end{minipage}

\subsection{Correction Behavior versus Tool-Augmented Baselines}
\noindent
Under shared initial responses and fixed denominators, Figure~\ref{fig:baseline_transition_flow} decomposes the final results of ProgCo, ProgCo-Py, and BiVerify into four state transitions. All three methods face the same 543 initially correct and 647 initially incorrect instances, making their transition rates directly comparable.

BiVerify achieves a \(W\!\rightarrow\!C\) correction rate of \(40.3\%\), exceeding ProgCo's \(9.4\%\) and ProgCo-Py's \(12.2\%\) by \(30.9\) and \(28.1\) percentage points, respectively. Meanwhile, their \(C\!\rightarrow\!W\) damage rates are \(3.7\%\), \(4.2\%\), and \(4.1\%\), indicating that BiVerify does not materially increase damage to initially correct answers while correcting more errors. In absolute counts, ProgCo, ProgCo-Py, and BiVerify correct 61, 79, and 261 initially incorrect instances and damage 20, 23, and 22 initially correct instances, yielding net correction counts of 41, 56, and 239. BiVerify's primary advantage over the programmatic-verification baselines therefore comes from broader error coverage rather than more aggressive indiscriminate rewriting.
\par\smallskip
\noindent\begin{minipage}{\textwidth}
\centering
\includegraphics[width=0.7\textwidth]{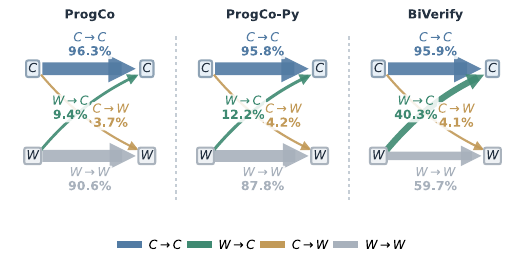}
\captionof{figure}{Final correctness state flows for ProgCo, ProgCo-Py, and BiVerify. Arrow colors denote the four transitions \(C\!\rightarrow\!C\), \(W\!\rightarrow\!C\), \(C\!\rightarrow\!W\), and \(W\!\rightarrow\!W\); arrow widths correspond to transition rates normalized by initial state. All three methods are evaluated on the same 1,190 model--problem instances.}
\label{fig:baseline_transition_flow}
\end{minipage}

\subsection{Correction Retention and Trajectory Stability}
\noindent
Figure~\ref{fig:trajectory_classes} presents the complete classification of multi-event correctness trajectories. By retaining intermediate event states, it distinguishes behaviors that final \(C/W\) transitions alone cannot identify, including stable repairs, fluctuations after repair, lost temporary repairs, and recovery after damage.

Among the 261 final \(W\rightarrow C\) instances, 248 remain correct after first becoming correct, while 13 fluctuate during the process but are correct at the end; stable repairs thus account for \(95.0\%\). Another 27 instances are temporarily corrected but ultimately become incorrect again, and 16 recover after being damaged at an intermediate event. Including temporary corrections, 288 initially incorrect instances become correct after at least one event, of which 261 remain correct at the end, corresponding to a final retention rate of \(90.6\%\). The overall gain therefore comes primarily from corrections retained to the end, although later events can introduce limited fluctuations.
\par\smallskip
\noindent\begin{minipage}{\textwidth}
\centering
\includegraphics[width=0.44\textwidth]{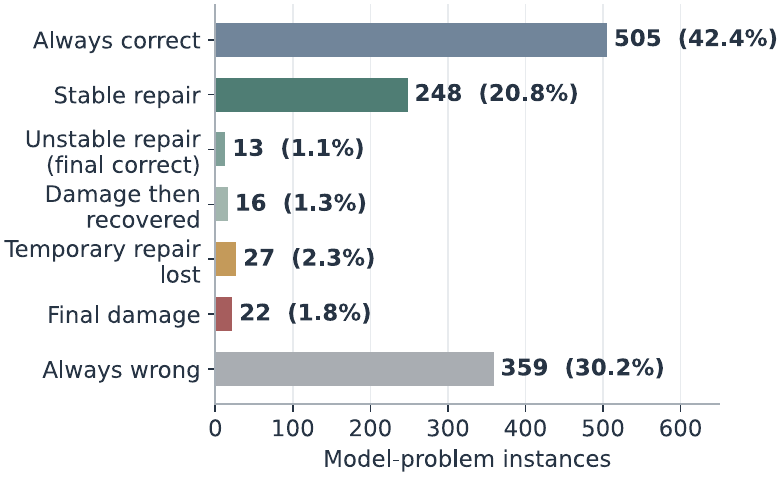}
\captionof{figure}{Multi-event correctness trajectory classes for BiVerify. Stable repair denotes an initially incorrect instance that remains correct after first becoming correct; Temporary repair lost denotes an instance that is temporarily corrected but becomes incorrect again by the final state.}
\label{fig:trajectory_classes}
\end{minipage}

\subsection{Actual Scheduling Paths}
\noindent
\begin{minipage}[t]{0.485\textwidth}
\vspace{0pt}
Figure~\ref{fig:path_analysis} groups instances into eight scheduling paths according to their actual B/F sequences. B\(\rightarrow\)F is the most common path, containing 421 instances with both initial and final accuracy of approximately \(81.0\%\). By contrast, paths such as B\(\rightarrow\)B\(\rightarrow\)F, B\(\rightarrow\)B\(\rightarrow\)B\(\rightarrow\)F, and B\(\rightarrow\)B\(\rightarrow\)F\(\rightarrow\)F begin with lower accuracy but improve by approximately \(37.0\), \(47.2\), and \(46.9\) percentage points, respectively. Although the path with four consecutive backward-verification events improves by \(17.3\) percentage points, its final accuracy remains only \(30.1\%\), reflecting greater residual correction difficulty in this group.

Because the scheduler selects these paths according to runtime states, differences among paths are descriptive rather than causal effects under randomized control. Nevertheless, the path distribution shows that BiVerify can terminate samples that reach consistent judgments early while allocating more verification events to samples that retain failure signals.
\end{minipage}\hfill
\begin{minipage}[t]{0.485\textwidth}
\vspace{0pt}
\centering
\includegraphics[width=0.8\linewidth]{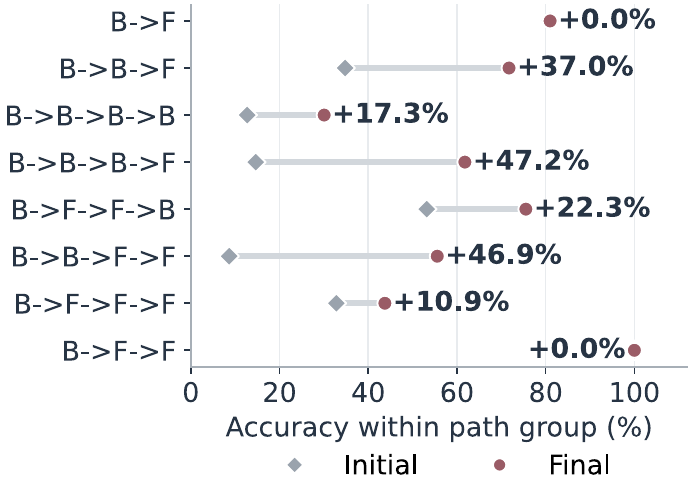}
\captionof{figure}{Initial and final accuracy for different actual scheduling paths in BiVerify. B and F denote backward and forward verification, respectively. Diamonds and circles show initial and final accuracy within each path group, and the percentage beside each circle indicates the absolute accuracy change from initial to final.}
\label{fig:path_analysis}
\end{minipage}

\subsection{Cross-Model Event Trajectories}
\noindent
Figure~\ref{fig:cross_model_trajectory} shows event-by-event accuracy across the seven base-model settings. From Initial to Final, BiVerify improves by \(27.1\), \(37.1\), \(47.7\), \(15.9\), \(4.7\), and \(8.3\) percentage points on DeepSeek-Flash, DeepSeek-Pro, GPT-5.4-Mini, GPT-5.4-Mini-Think, Gemini-2.5-Flash-Lite, and Gemini-2.5-Flash-Lite-Think, respectively. DeepSeek-Flash-Think starts at \(88.8\%\) and remains at \(88.8\%\) at the end. Intermediate trajectories are not uniformly monotonic across settings; for example, some models exhibit slight fluctuations in later events. Thus, the gains from bidirectional verification depend on the base model's initial capability and remaining correction headroom, while later events can still introduce a risk of over-correction.
\par\smallskip
\noindent
\begin{minipage}{\textwidth}
\centering
\includegraphics[width=0.7\textwidth]{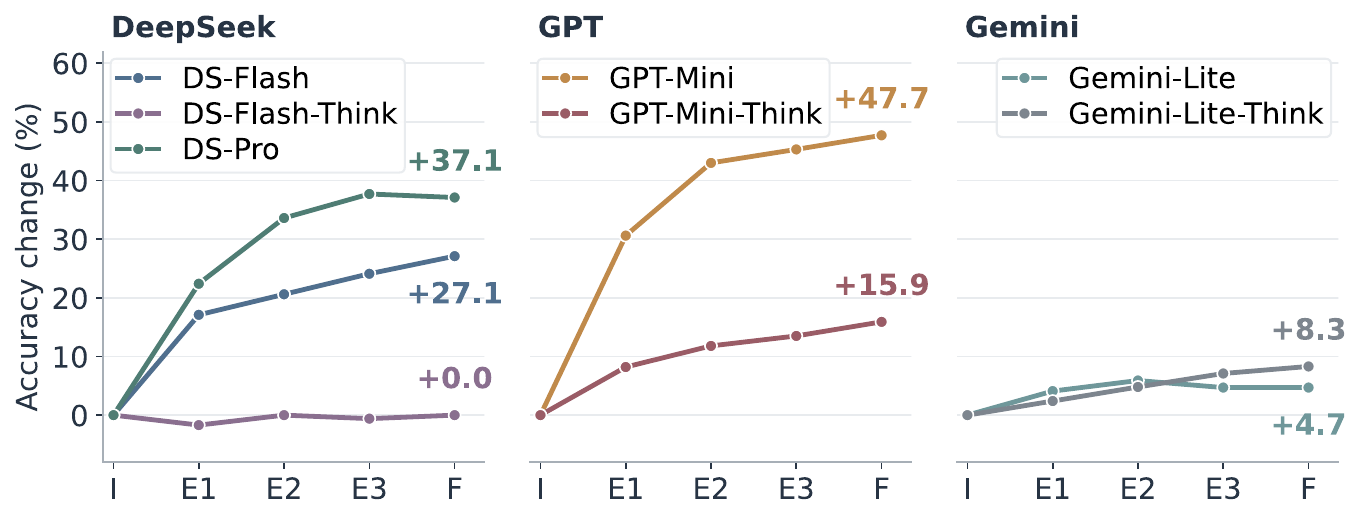}
\captionof{figure}{Event-by-event accuracy trajectories of BiVerify across seven base-model settings. The three panels correspond to the DeepSeek, GPT, and Gemini model families, and the vertical axis shows absolute change from each model's Initial accuracy. I, E1--E3, and F denote Initial, the first three actual verification-event positions, and Final, respectively. Final aggregates the result after at most four actual verification events, and the percentage at each curve endpoint indicates the absolute accuracy change from Initial to Final.}
\label{fig:cross_model_trajectory}
\end{minipage}

\FloatBarrier
 \section{Complete Pseudocode for BiVerify}

\par\medskip
\refstepcounter{algorithm}
\noindent\rule{\textwidth}{0.8pt}
\par\noindent Algorithm~\thealgorithm: BiVerify
\label{alg:biverify}
\par\noindent\rule{\textwidth}{0.4pt}
{\small
\begin{algorithmic}[1]
\REQUIRE Problem \(q\), initial solution \(y_0\), maximum number of actual verification events \(T\)
\ENSURE Final candidate \(C\), joint-verification indicator \(z\)

\STATE \(a\leftarrow\operatorname{Extract}(y_0)\)
\STATE \(C\leftarrow(q,y_0,a)\)
\STATE \(B\leftarrow\mathrm{Unknown},\;
       F\leftarrow\mathrm{Unknown}\)
\STATE \(d\leftarrow\mathrm{B}\)
\STATE Initialize or construct \(V^{B},K,V^{F}\) on demand using \(\mathcal A_{\mathrm{flow}}^{B}\) and \(\mathcal A_{\mathrm{flow}}^{F}\)
\STATE Each agent may invoke the shared mathematical tool agent \(\mathcal A_{\mathrm{tool}}\) as needed

\FOR{\(t=0\) to \(T-1\)}
    \IF{\(B=\mathrm{Pass}\ \land\ F=\mathrm{Pass}\)}
        \STATE \RETURN \(C,1\)
    \ENDIF

    \IF{\(d=\mathrm{B}\)}
        \STATE \((s,E)\leftarrow
        \mathcal A_{\mathrm{ver}}^{B}(C,V^{B})\)
    \ELSE
        \STATE \((s,E,k)\leftarrow
        \mathcal A_{\mathrm{ver}}^{F}(C,K,V^{F})\)
    \ENDIF

    \IF{\(s=\mathrm{Pass}\)}
        \IF{\(d=\mathrm{B}\)}
            \STATE \(B\leftarrow\mathrm{Pass}\)
        \ELSE
            \STATE \(F\leftarrow\mathrm{Pass}\)
        \ENDIF

        \IF{\(B=\mathrm{Pass}\ \land\ F=\mathrm{Pass}\)}
            \STATE \RETURN \(C,1\)
        \ELSE
            \STATE \(d\leftarrow\operatorname{Other}(d)\)
        \ENDIF

    \ELSE
        \STATE \(\widehat C\leftarrow
        \mathcal A_{\mathrm{update}}(C,E,d)\)

\IF{\(\widehat C\) is adopted by scheduler \(\mathcal S\) as a new candidate}
            \STATE \(C\leftarrow\widehat C\)
            \STATE \(B\leftarrow\mathrm{Unknown},\;
                   F\leftarrow\mathrm{Unknown}\)
\STATE Invalidate the previous verification flows and regenerate them for \(C\) on demand
\STATE Retain the current direction \(d\)
        \ELSE
\STATE \(\mathcal A_{\mathrm{update}}\) revises the verification flow in the current direction using \(C,E,d\)
\STATE Retain the current direction \(d\)
        \ENDIF
    \ENDIF
\ENDFOR

\STATE \RETURN \(C,0\)
\end{algorithmic}
}
\noindent\rule{\textwidth}{0.8pt}

For notational simplicity, the algorithm omits the event subscript \(t\). Here, \(s\), \(E\), and \(k\) denote the verdict in the current direction, verification evidence, and the position of the first failed claim, respectively. The pseudocode directly uses the verification-flow agents, adjudication agents, verification-state and candidate-update agent, and shared mathematical tool agent shown in the framework figure, without expanding their internal prompts or tool-call details.
 \end{document}